%% file: sample_FG2019.tex
\documentclass[a4paper, 10pt, conference]{ieeeconf}      
\usepackage{FG2019}
\usepackage{xspace}
\usepackage{graphicx}
\usepackage{amsmath,amssymb} 
\usepackage{color}
\usepackage{booktabs}
\usepackage{lipsum}
\usepackage{xcolor}
\usepackage{multirow}
\usepackage{wrapfig}
\usepackage{tikz}
\usepackage{setspace}
\usepackage{pgfplots}
\pgfplotsset{compat=newest}
\usepackage{balance}
\usepackage{xcolor}

\FGfinalcopy 

\IEEEoverridecommandlockouts                              
\def\FGPaperID{4} 

\title{\LARGE \bf Self-Supervised Learning of Face Representations for \\
Video Face Clustering}

\author{\parbox{16cm}{\centering
    {\large Vivek Sharma$^1$, Makarand Tapaswi$^2$, M. Saquib Sarfraz$^1$ and Rainer Stiefelhagen$^1$}\\
    {\normalsize
    $^1$ CV:HCI, Karlsruhe Institute of Technology, Karlsruhe, Germany\\
    $^2$ University of Toronto, Toronto, Canada}}
}

\begin{document}
\makeatletter
\DeclareRobustCommand\onedot{\futurelet\@let@token\@onedot}
\def\@onedot{\ifx\@let@token.\else.\null\fi\xspace}

\def\eg{\emph{e.g}\onedot} \def\Eg{\emph{E.g}\onedot}
\def\ie{\emph{i.e}\onedot} \def\Ie{\emph{I.e}\onedot}
\def\cf{\emph{c.f}\onedot} \def\Cf{\emph{C.f}\onedot}
\def\etc{\emph{etc}\onedot} \def\vs{\emph{vs}\onedot}
\def\wrt{w.r.t\onedot} \def\dof{d.o.f\onedot}
\def\etal{\emph{et al}\onedot}
\makeatother

\def\bx{\mathbf{x}}
\def\bt{\mathbf{t}}

\ifFGfinal
\thispagestyle{empty}
\pagestyle{empty}
\else
\author{Anonymous FG 2019 submission\\ Paper ID \FGPaperID \\}
\pagestyle{plain}
\fi
\maketitle

\begin{abstract}
Analyzing the story behind TV series and movies often requires understanding who the characters are and what they are doing.
With improving deep face models, this may seem like a solved problem.
However, as face detectors get better, clustering/identification needs to be revisited to address increasing diversity in facial appearance.
In this paper, we address video face clustering using unsupervised methods.
Our emphasis is on distilling the essential information, \emph{identity}, from the representations obtained using deep pre-trained face networks.
We propose a self-supervised Siamese network that can be trained without the need for video/track based supervision, and thus can also be applied to image collections.
We evaluate our proposed method on three video face clustering datasets. The experiments show that our methods outperform current state-of-the-art methods on all datasets.
Video face clustering is lacking a common benchmark as current works are often evaluated with different metrics and/or different sets of face tracks.
The datasets and code are available at \textcolor{blue}{https://github.com/vivoutlaw/SSIAM}.

\end{abstract}

\input{intro.tex}

\input{relwork.tex}
\input{model.tex}
\input{experiments.tex}

\input{conclusion.tex}

\noindent
\textbf{Acknowledgements:} This work is supported by the DFG, a German Research Foundation - funded PLUMCOT project.

{
\bibliographystyle{ieee}
\bibliography{sample_FG2019}
}

\end{document}

%% file: intro.tex
\section{Introduction}
\label{sec:intro}

Large videos such as TV series episodes or movies undergo several preprocessing steps to make the video more accessible, \eg~shot detection. Character identification/clustering has become one such important step with several emerging research areas~\cite{lsmdc,movieqa,moviegraphs} requiring it. For example, in video question-answering~\cite{movieqa}, most questions center around the characters asking who they are, what they do, and even why they act in certain ways. The related task of video captioning~\cite{lsmdc} often uses a character agnostic way (replacing names by \emph{someone}) making the captions very artificial and uninformative (\eg~\emph{someone} opens the door). However, recent work~\cite{rohrbach2017groundedpeople} suggests that more meaningful captions can be achieved from an improved understanding of characters. In general, the ability to predict which characters appear when and where facilitates a deeper video understanding that is grounded in the storyline.

Motivated by this goal, person clustering~\cite{cinbis2011unsupervised,ocik,erdosclustering,jfac} and  identification~\cite{everingham2006,sivic2009,baeuml2013,ramanathan2014,nagrani2017} in videos has seen over a decade of research. In particular, fully automatic person identification is achieved either by aligning subtitles and transcripts~\cite{everingham2006,sivic2009,baeuml2013}, or using web images for actors and characters~\cite{aljundi2016breakbad,nagrani2017} as a form of weak supervision. On the other hand, clustering~\cite{jfac,cinbis2011unsupervised,wu2013simultaneous,wu2013constrained,imptriplet,bestfgr2018} has mainly relied on \emph{must-link} and \emph{cannot-link} information obtained by tracking faces in a shot and analyzing their co-occurrence.

As face detectors improve (\eg~\cite{tinyface}), clustering and identification need to be revisited as more faces that exhibit extreme viewpoints, illumination, resolution, become available and need to be grouped or identified. Along with improvements to face detection, deep Convolutional Neural Networks (CNNs) have also yielded large performance gains for face representations~\cite{deepface,facenet,vggface,vggface2}. These networks are typically trained using hundreds-of-thousands to millions of face images gathered from the web, and show super-human performance on face verification tasks on images (LFW~\cite{lfw}) and videos (YouTubeFaces~\cite{youtubefaces}). Nevertheless, it is important to note that faces in videos such as TV series/movies exhibit more variety in comparison to \eg~LFW, where the images are obtained from Yahoo News by cropping mostly frontal faces. While these deep models generalize well, they are difficult to train from scratch (require lots of training data), and are typically transferred to other datasets via \emph{net surgery}: fine-tuning~\cite{die,imptriplet,jfac}, or use of additional embeddings on the features from the last layer~\cite{sarfraz2017pose,tle}, or both.

Recent work shows that CNN representations can be improved via positive and negative pairs that are discovered through an MRF~\cite{jfac}; or a revised triplet-loss~\cite{imptriplet}.   In contrast, we propose simple methods that  do not require complex optimization functions or supervision to improve the feature representation. We emphasize that  while video-level constraints aren’t new, they need to be used properly to extract the most out of them. This is especially in light of CNN face representations that are very similar even across different identities. Fig.~\ref{fig:dist_histogram} proves this point as we see a large overlap between the cosine similarity distributions of positive (same id) and negative (across id) track pairs on the base features.

In this paper, we focus on the video face clustering problem. Given a set of face tracks from several characters, our goal is to group them so that tracks in a cluster belong to the same character. We summarize the main contributions of our paper, and also highlight key differences:
(1) We propose two variants of discriminative methods (Sec.~\ref{sec:model}) that build upon deep network representations, to learn discriminative facial representations. They are Track-supervised Siamese network (TSiam) and Self-supervised Siamese network (SSiam). In contrast to previous methods~\cite{bestfgr2018}, in TSiam, we incorporate negative pairs for the singleton (no co-occurring) tracks. In SSiam, we obtain positives and negatives by sorting distances (\ie~ranking) on a subset of frames.  Note that methods proposed in this paper are either fully unsupervised, or use supervision that is obtained automatically, hence can be thought as unsupervised. Additionally, our fully unsupervised method can mine positive and negative pairs without the need for tracking. This enables application of our method to image collections.
(2) We perform extensive empirical studies and demonstrate the effectiveness and generalisation of the methods. Our methods are powerful and obtain performance comparable or higher than state-of-the-art when evaluated on three challenging video face clustering datasets.

%% file: relwork.tex
\section{Related Work}
\label{sec:relatedwork}

Over the last decade, video face clustering is typically modeled using discriminative methods to improve face representations. In the following, we review some related work in this area.

\vspace{2mm}
\noindent\textbf{Video face clustering.}
Clustering faces in videos commonly uses pairwise constraints obtained by analyzing tracks and some form of representation/metric learning.
Face image pairs belonging to the same track are labeled positive (same character), while face images from co-occurring tracks help create negatives (different characters).
This strategy has been exploited by learning a metric to obtain cast-specific distances~\cite{cinbis2011unsupervised} (ULDML);
iteratively clustering and associating short sequences based on hidden Markov Random Field (HMRF)~\cite{wu2013simultaneous,wu2013constrained};
or performing clustering in a sub-space obtained by a weighted block-sparse low-rank representation (WBSLRR)~\cite{xiao2014weighted}.
In addition to pairwise constraints, video editing cues are used in an unsupervised way to merge tracks~\cite{tc}.
Here, track and cluster representations are learned on-the-fly with dense-SIFT Fisher vectors~\cite{vf2}.
Recently, Jin~\etal~\cite{erdosclustering} consider detection and clustering jointly, and propose a link-based clustering (Erd\"{o}s-R\'{e}nyi) based on rank-1 counts verification.
The linking is done by comparing a given frame with a reference frame and a threshold is learned to merge/not-merge frames.

Face track clustering/identification methods have also used additional cues such as
clothing appearance~\cite{tapaswi2012}, speech~\cite{paul2014conditional}, voice models~\cite{nagrani2017}, context~\cite{zhang2013unified}, gender~\cite{mcafc}, name mentions (first, second, and third person references) in subtitles~\cite{Haurilet2016}, weak labels using transcripts/subtitles~\cite{everingham2006,baeuml2013}, and joint action and actor labeling~\cite{miech2017learning} using transcripts.

With the popularity of CNNs, there is a growing focus on improving face representations using video-level constraints. An improved form of triplet loss is used to fine-tune the network and push the positive and negative samples apart in addition to requiring anchor and positive to be close, and anchor and negative far~\cite{imptriplet}. Zhang~\etal~\cite{jfac} learn better representations by dynamic clustering constraints that are discovered iteratively during clustering that is performed via a Markov Random Field (MRF). In contrast to related work, we propose a simple, yet effective approach (SSiam) to learn good representations by sorting distances on a subset of frames and not requiring video/track level constraints to generate positive/negative training pairs. 

Further,  Zhang~\etal~\cite{imptriplet}-\cite{jfac} and Datta~\etal~\cite{bestfgr2018} utilize video-level constraints only to generate a set of similar and dissimilar face pairs. Thus, the model does not see negative pairs for singleton (no co-occurring) tracks. In contrast, our method TSiam incorporates negative pairs for the singleton tracks by exploiting track-level distance.

Finally, it is worth noting the work on ``harvesting" training data from unlabeled sources which is in the similar spirit of SSiam and TSiam.  Fernando~\etal~\cite{fernando2017self} and Mishra~\etal~\cite{ref2} shuffle the video frames and treat them as positive/negative training data for reordering video frames; Wang~\etal~\cite{ref3} collect positive/negative training data by tracking bounding boxes (\ie~motion information) in order to learn effective visual representations.

%% file: model.tex
\section{Refining Face Representations for Clustering}
\label{sec:model}

Our goal is to improve face representations using simple methods that build upon the success of deep CNNs.
More precisely, we propose models to refine the face descriptors automatically, without the need for manual labels.
In contrast, fine-tuning the original CNN typically requires supervised class labels.
Thus, our approach has three key benefits:
(i) it is easily applicable to new videos;
(ii) it does not need large amounts of training data (few hundred tracks are enough); and
(iii) specialized networks can be trained on each episode, film, or series.

\begin{figure}[t]
\centering
\includegraphics[width=0.99\columnwidth]{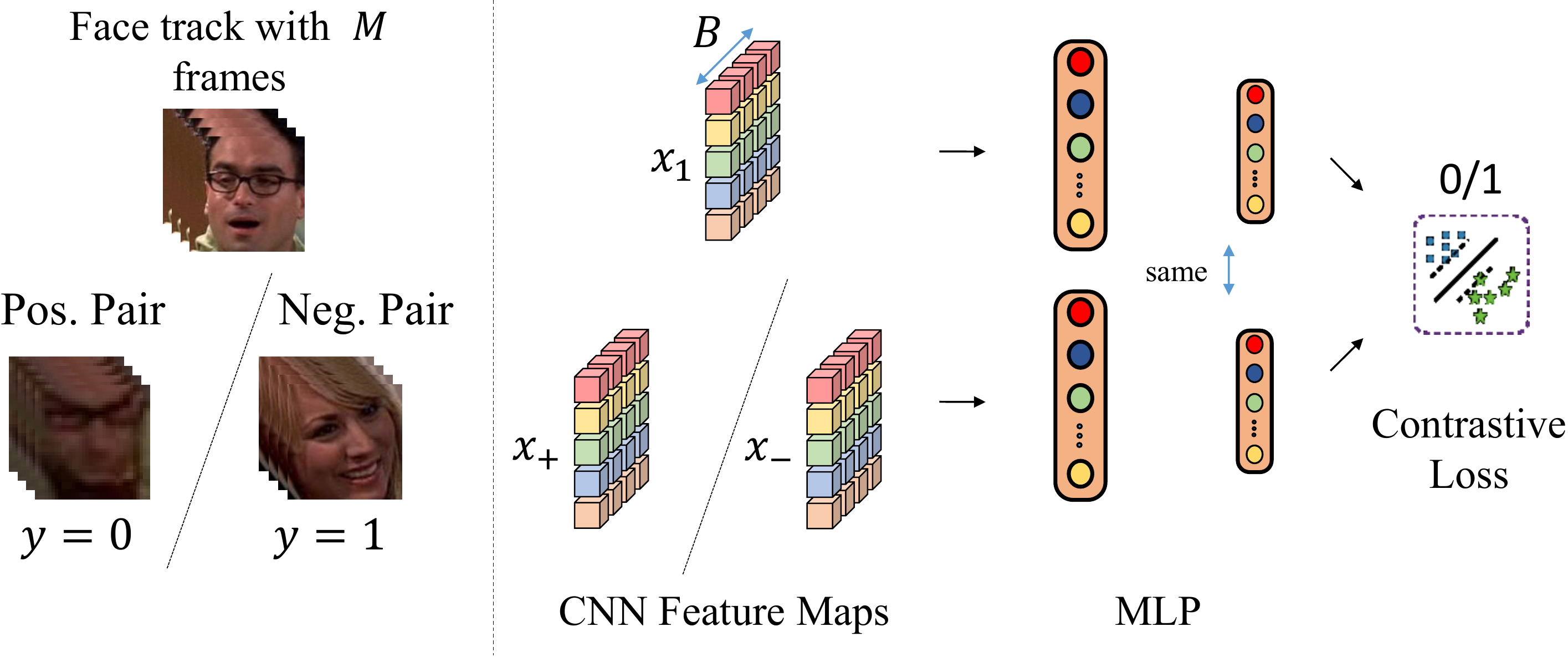}
\caption{\textbf{Track-supervised Siamese network (TSiam)}. Illustration of the Siamese architecture used in our track-supervised Siamese networks.
Note that the MLP is shared across both feature maps.
$B$ corresponds to batch size.} 
\label{fig:tsiam}
\end{figure}

\begin{figure*}[t]
\centering
\includegraphics[width=1.9\columnwidth]{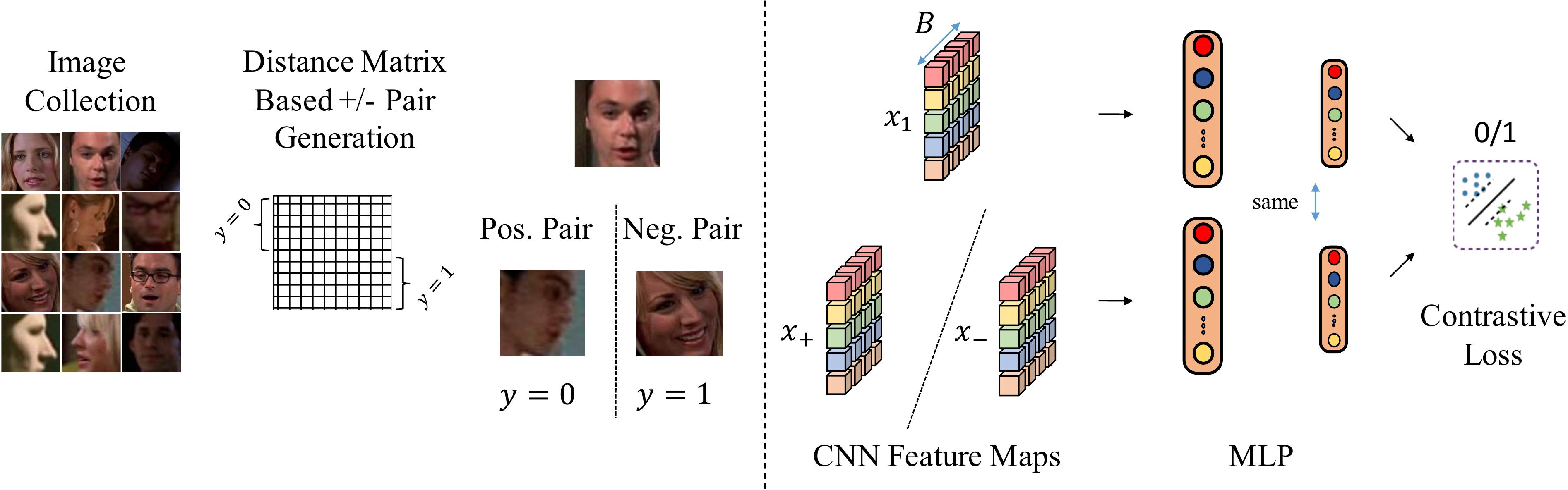}
\caption{\textbf{Self-supervised Siamese network (SSiam)}. Illustration of the Siamese architecture used in our self-supervised Siamese networks.
SSiam selects hard pairs: farthest positives and closest negatives using a ranked list based on Euclidean distance for learning similarity and dissimilarity respectively. 
Note that the MLP is the same across both feature maps. $B$ corresponds to batch.} 
\label{fig:ssiam}
\end{figure*}

\vspace{2mm}
\noindent\textbf{Preliminaries.}
Consider a video with $N$ face tracks $\{T^1, \ldots, T^N\}$ belonging to $C$ characters.
Each track corresponds to one of the characters, and consists of $T^i = \{f_1, \ldots, f_{M^i}\}$ face images.
Our goal is to group tracks into $\{G_1, \ldots, G_{|C|}\}$ such that each track is assigned to only one group, and ideally, each group contains all tracks from the same character.
We use a deep CNN (VGG2~\cite{vggface2}) and extract a descriptor for each face image $\bx^i_k \in \mathbb{R}^D, \, k = 1, \ldots, M^i$ from the penultimate layer (before classification) of the network.
We refer to these as base features, and demonstrate that they already achieve a high performance.
As a form of data augmentation, we use 10 crops obtained from an expanded bounding box surrounding the face image during training.
Evaluation is based on one center crop.

Track-level representations are obtained by aggregating the face image descriptors
\begin{equation}
\bt^i = \frac{1}{M^i} \sum_k \bx^i_k \, .
\end{equation}
We additionally normalize track representations to be unit-norm, $\hat{\bt}^i = \bt^i / \|\bt^i\|_2$ before using them for clustering.

In~\cite{tc,imptriplet,jfac}, the authors use Hierarchical Agglomerative Clustering (HAC) as the clustering technique. For a fair comparison to \cite{tc,imptriplet,jfac}, we perform HAC to obtain a fixed number of clusters, equal to the number of characters.
We use the minimum variance ward linkage~\cite{ward_linkage} for all methods. See Fig.~\ref{fig:fg_clust} for illustration.

\vspace{2mm}
\noindent\textbf{Discriminative models.}
Discriminative clustering models typically associate a label $y$ with a pair of features.
We choose $y = 0$ when a pair of features $(\bx_1, \bx_2)$ should belong to the same cluster, and $y = 1$ otherwise~\cite{contrastive_loss}.

We use a shallow MLP to reduce the dimensionality and improve generalization of the features (see Fig.~\ref{fig:tsiam},~\ref{fig:ssiam}).
Here, each face image is encoded as $Q_\phi(\bx^i_k)$, where $\phi$ corresponds to the trainable parameters.
We find $Q_\phi(\cdot)$ to perform best when using a linear layer (for details see Sec.~\ref{subsec:implementation}).
To perform clustering, we compute track-level aggregated features by average pooling across the embedded frame-level representations~\cite{sharma2017}
\begin{equation}
\bt^i = \frac{1}{M^i} \sum_k Q_\phi(\bx^i_k) \, ,
\end{equation}
followed by $\ell_2$-normalization.

We train our model parameters at the frame-level by minimizing the contrastive loss~\cite{contrastive_loss}:
\begin{equation}
\label{eq:contrasiveloss}
\begin{split}
\mathcal{L} \left(W, y, Q_\phi(\bx_1), Q_\phi(\bx_2) \right) =  \qquad \qquad \qquad \qquad \qquad \\
\frac{1}{2} \left( (1 - y) \cdot (d_W)^2 + y \cdot (\max(0, m - d_W))^2 \right) \, ,
\end{split}
\end{equation}
where $\bx_1$ and $\bx_2$ are a pair of face representations with $y = 0$ when coming from the same character, and $y = 1$ otherwise.
$W: \mathbb{R}^{D \times d}$ is a linear layer that embeds $Q_\phi(\bx)$ such that $d \ll D$ (in our case, $d = 2$).
$d_W$ is the Euclidean distance $d_W = \| W \cdot Q_\phi(\bx_1) - W \cdot Q_\phi(\bx_2) \|^2$, and
$m$ is the margin, empirically chosen to be 1.

In the following, we present two strategies to automatically obtain supervision for pairs of frames:  Fig.~\ref{fig:tsiam} illustrates  the Track-level supervision, and Fig.~\ref{fig:ssiam} shows the  Self-supervision for Siamese network training.

\vspace{2mm}
\noindent\textbf{Track-supervised Siamese network (TSiam).} 
Video face clustering often employs face tracking to group face detections made in each frame.
The tracking acts as a form of high precision clustering (grouping detections within a shot) and is popularly used to automatically generate positive and negative pairs of face images~\cite{cinbis2011unsupervised,bestfgr2018,tc,wu2013constrained}. 
In each frame, we assume that characters appear on screen only once.
Thus, all face images within a track can be used as positive pairs, while face images from co-occurring tracks are used as negative pairs.
For each frame in the track, we sample two frames within the same track to form positive pairs, and sample four frames from a co-occurring track (if it exists) to form negative pairs. 

Depending on the filming style of the series/movie, characters may appear alone or together on screen.
As we perform experiments on diverse datasets, for some videos 35\% tracks have co-occurring tracks, while this can be as large as 70\% for other videos.
For isolated tracks, we sort all other tracks based on track-level distances (computed on base features) and randomly sample frames from the farthest $F = 25$ tracks. Note that all previous works ignore negative pairs for singleton (not co-occurring) tracks. We will highlight their impact in our experiments.

\vspace{2mm}
\noindent\textbf{Self-supervised Siamese network (SSiam).}
Clustering is an unsupervised task and supervision from tracks may not always be available.
An example is face clustering within image collections (\eg~on social media platforms).
To enable the use of metric learning without any supervision we propose an effective approach that can generate the required pairs automatically during training.
SSiam is inspired by pseudo-relevance feedback (pseudo-RF)~\cite{psuedo1,pseudo2} that is commonly used in information retrieval.

We hypothesize that the first and last samples of a ranked list based on Euclidean distance are strong candidates for learning similarity and dissimilarity respectively.
We exploit this in a meaningful way and generate promising similar and dissimilar pairs from a representative subset of the data.

Formally, consider a subset $\mathcal{S} = \{\bx_1, \ldots, \bx_B\}$ of frames from the dataset.
We treat each frame $\bx_b, b = 1,\ldots,B$ as a query and compute Euclidean distance against every other frame in the set.
We sort rows of the resulting matrix in an ascending order (smallest to largest distance) to obtain an ordered index matrix $\mathcal{O}(\mathcal{S}) = [ s_1^o; \ldots; s_B^o ]$.
Each row $s_b^o$ contains an ordered index of the closest to farthest faces corresponding to $\bx_b$.
Note that the first column of such a matrix is the index itself at distance 0.
The second column corresponds to nearest neighbors for each frame and can be used to form the set of positive pairs $\mathcal{S}_+$.
Similarly, the last column corresponds to farthest neighbors and forms the set of negative pairs $\mathcal{S}_-$.
Each element of the above sets stores: query index $b$, nearest/farthest neighbor $r$, and the Euc. distance $d$.
However, we still need to consider the choice of:
(i) the set of frames, and
(ii) training pairs from the candidate sets of positive and negative pairs. 

We address this during training and form the pairs dynamically by picking a random subset of $B$ frames at each iteration.
We compute the distances, sort them, and obtain positive and negative pairs sets $\mathcal{S}_+, \mathcal{S}_-$, each with $B$ elements.
Among them, we choose $K$ pairs from the positive set that have the largest distances and $K$ pairs from the negative set with the smallest distances.
This allows us to select pairs semi-hard positives and semi-hard negatives from the representative set of $B$ frames.
Finally, these $2K$ pairs form the training batch for the network.  

To encourage variety in the sample set $\mathcal{S}$ and reduce the chance of false positives/negatives in the chosen $2K$ pairs, $B$ is chosen to be much larger than $K$ ($B = 1000, K = 64$). Experiments on difficult datasets and generalization studies show the benefit and effectiveness of this approach in collecting positive and negative pairs to train the network. 

Note that, SSiam can be thought of as an improved version of pseudo-RF with batch processing. Rather than selecting farthest negatives and closest positives for each independent query, we emphasize that SSiam selects 2$K$ hard pairs: farthest positives and closest negatives by looking at the batch of queries $B$ jointly. This selection of sorted pairs from the positive $\mathcal{S}_+$ and negative $\mathcal{S}_-$ sets is quite important. See experiments for a more detailed comparison of SSiam with pseudo-RF.

%% file: experiments.tex
\section{Evaluation}
\label{sec:eval}

\begin{figure}[t]
\centering
\includegraphics[width=0.99\columnwidth]{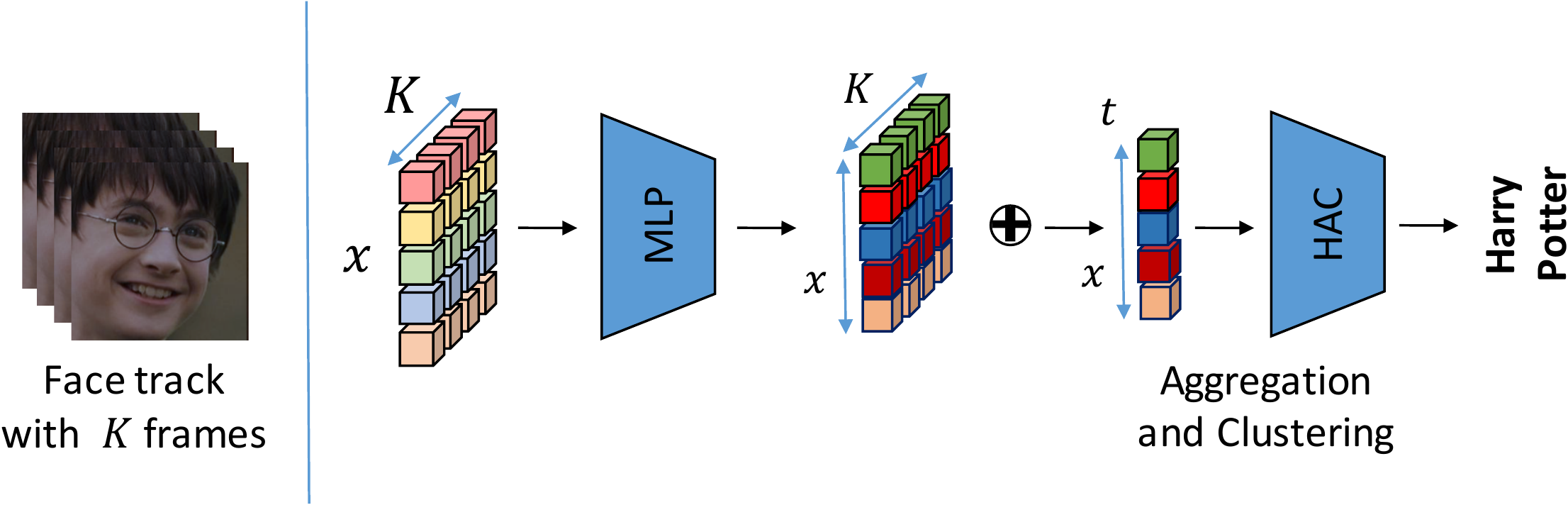}
\caption{\textbf{Illustration of the testing evaluation}. Given our pre-trained MLP models: TSiam and SSiam. We extract the frame-level features for the track, followed by mean pool for a track-level representation, which is then fed to HAC clustering with fixed number of clusters. Finally, we evaluate the quality of clustering via Accuracy or BCubed F-measure.} 
\label{fig:fg_clust}
\end{figure}

We present our evaluation on three challenging datasets.
We first describe the clustering metric, followed by a thorough analysis of the proposed methods, ending with a comparison to state-of-the-art.

\subsection{Experimental Setup}

\noindent\textbf{Datasets.}
We conduct experiments on three challenging video face identification/clustering datasets:
(i) \textit{Buffy the Vampire Slayer} (BF)~\cite{baeuml2013,jfac} (season 5, episodes 1 to 6): a drama series with several shots in the dark at night;
(ii) \textit{Big Bang Theory} (BBT)~\cite{tapaswi2012,baeuml2013,wu2013simultaneous,imptriplet} (season 1, episodes 1 to 6): a sitcom with small cast list shot mainly indoors, and (iii) \textit{ACCIO}~\cite{accio}: \textit{Accio-1} first installment of ``\textit{Harry Potter}'' movie series with a large number of dark scenes and several tracks with non-frontal faces.

\begin{table}[h]
\small
\tabcolsep=1.2mm
\begin{center}
\caption{Dataset statistics for BBT~\cite{wu2013simultaneous,wu2013constrained,imptriplet}, BF~\cite{jfac} and ACCIO~\cite{jfac}.} 
\label{table:stats}
\begin{tabular}{lcccc}
\toprule
				&		 &   This work  &               & Previous work \\
Datasets        & \#Cast &  \#TR (\#FR) &    LC/SC (\%) & \#TR (\#FR)\\ 
\midrule
BBT0101 		& 	5	&	644 (41220)	&	37.2 / 4.1  & 182 (11525)\\
BF0502 			&	6	&	568 (39263)	&	36.2 / 5.0  & 229 (17337)\\
ACCIO	   		&	36	&	3243 (166885) & 30.93/0.05&	3243 (166885) \\
\bottomrule
\end{tabular}
\end{center}
\end{table}

\begin{figure*}[t]
\centering
\includegraphics[width=1.9\columnwidth]{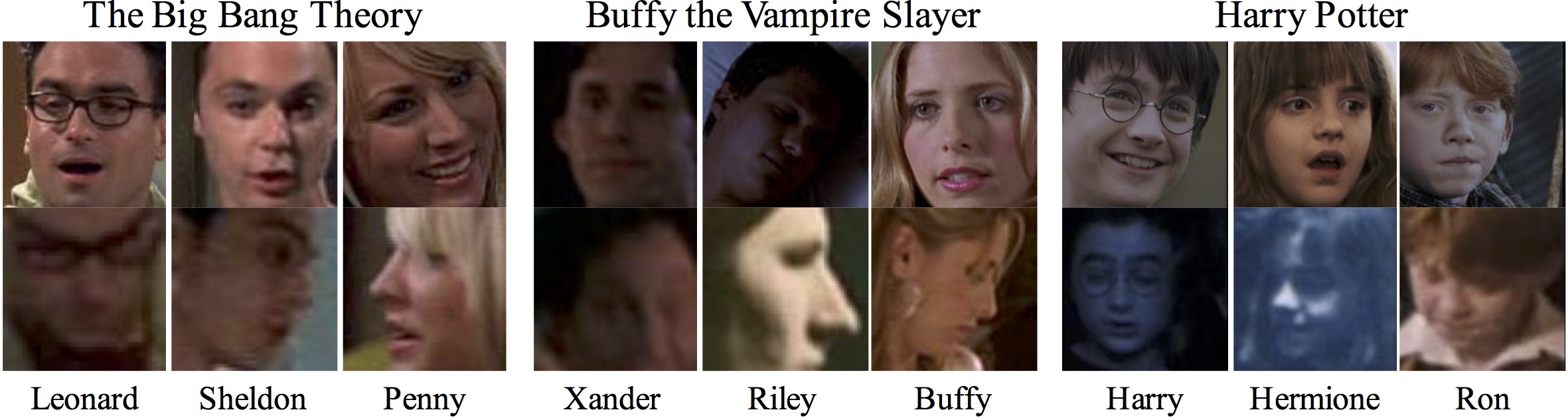}
\caption{Example images for a few characters from our dataset.
We show one easy view and one difficult view.
The extreme variation in illumination, pose, resolution, and attributes (spectacles) make the datasets challenging.}
\label{fig:dataset}
\end{figure*}

Most current works~\cite{cinbis2011unsupervised,wu2013constrained,imptriplet,jfac} that focus on improving feature representations for video-face clustering assume the number of main characters/clusters is known. In this work, for a fair comparison we follow the same protocols that are widely employed in the previous works~\cite{sharma2017,imptriplet,jfac} and train on BF episode 2, BBT episode 1, and ACCIO.
We also use the same number of characters as previous methods~\cite{jfac,imptriplet}, however, it is important to note that we do not discard tracks/faces that are small or have large pose variation. When not mentioned otherwise, we use an updated version of face tracks released by~\cite{baeuml2013} that incorporate several detectors to encompass all pan angles and in-plane rotations up to 45 degrees. Tracks are created via an online tracking-by-detection scheme with a particle filter.

We present a summary of our dataset in Table~\ref{table:stats}, and also indicate the number of tracks (\#TR) and frames (\#FR)  used in other works, showing that our data is indeed more challenging.
Additionally, it is important to note that different characters have wide variations in number of tracks, indicated by the class balance largest class (LC) to smallest class (SC). Fig.~\ref{fig:dataset} shows a few examples of difficult faces our methods have to deal with. 

\vspace{2mm}
\noindent\textbf{Evaluation metric.}
We use Clustering Accuracy (ACC)~\cite{jfac} also called Weighted Clustering Purity (WCP)~\cite{tc}  as the metric to evaluate the quality of clustering.
As we compare methods providing equal numbers of clusters (number of main cast), ACC is a fair metric for comparison.
ACC is computed as
\begin{equation}
\mbox{ACC} = \frac{1}{N} \sum_{c = 1}^{|C|} n_c \cdot p_c \, ,
\end{equation}
where $N$ is the total number of tracks in the video,
$n_c$ is the number of samples in the cluster $c$, and
cluster purity $p_c$ is measured as the fraction of the largest number of samples from the same label to $n_{c}$.
$|C|$ corresponds to the number of main cast members, and in our case also the number of clusters. For ACCIO, we report BCubed Precision (P), Recall (R) and F-measure (F) for a fair comparison with the state-of-the-art. Figure~\ref{fig:fg_clust} illustrates the steps for the  testing evaluation.

\subsection{Implementation Details}
\label{subsec:implementation}

\noindent\textbf{CNN.}
We adopt the VGG-2 face CNN~\cite{vggface2}, a ResNet50 model, pre-trained on MS-Celeb-1M~\cite{msceleb1m} and fine-tuned on 3.31M face images of 9131 subjects (VGG2 data).
Input RGB face images are resized to $224 \times 224$, and pushed through the CNN.
We extract \texttt{pool5\_7x7\_s1} features, resulting in $\bx^i_k \in \mathbb{R}^{2048}$.

\vspace{2mm}
\noindent\textbf{Siamese network MLP.}
The network comprises of fully-connected layers ($\mathbb{R}^{2048} \rightarrow \mathbb{R}^{256} \rightarrow \mathbb{R}^{2}$).
Note that the second linear layer is part of the contrastive loss (corresponds to $W$ in Eq.~\ref{eq:contrasiveloss}), and we use the feature representations at $\mathbb{R}^{256}$ for clustering.

To train our Siamese network with track-level supervision (TSiam), we obtain about 102k positive and 204k negative frame pairs (for BBT-0101) by mining 2 positive and 4 negative pairs for each frame.
For Self-supervised Siamese, we generate pairs as described in Sec.~\ref{sec:model}, using $B = 1000, K = 64$.
Higher values values of $B = 2000, 3000$, did not provide significant improvements.

The MLP is trained using the contrastive loss, and parameters are updated using Stochastic Gradient Descent (SGD) with a fixed learning rate of $10^{-3}$. Since the labels are obtained automatically, overfitting is not a concern. We train our model until convergence (loss does not reduce significantly).

\subsection{Clustering Performance and Generalization}

\begin{table}[h]
\small
\tabcolsep=0.2cm
\begin{center}
\caption{Clustering accuracy on the base face representations.}
\label{table:base_feats} 
\begin{tabular}{lcccc}
\toprule
\multirow{2}{*}{Dataset} &     \multicolumn{2}{c}{Track Level}  &     \multicolumn{2}{c}{Frame Level}    \\
					     & VGG1	& VGG2 & VGG1 & VGG2 \\
\midrule
BBT-0101 		&0.916	&\textbf{0.932}	& 0.938		& \textbf{0.940}	\\
BF-0502 		&0.831	&\textbf{0.836}	& 0.901		&\textbf{0.912} \\
\bottomrule
\end{tabular}
\end{center}
\end{table}

\noindent\textbf{Base features.}
We compare track- and frame-level performance of two popular deep face representations: VGG1~\cite{vggface} and VGG2~\cite{vggface2}. Track-level results use mean-pool of frames.
Number of clusters matches number of characters. Results are reported in Table~\ref{table:base_feats}.
Note that the differences are typically within 1\% of each other indicating that the results in the following tables are not just due to having better CNNs trained with more data.
We refer to VGG2 features as Base for the remainder of this paper.

\begin{figure}[t]
\centering
\includegraphics[width=0.75\columnwidth]{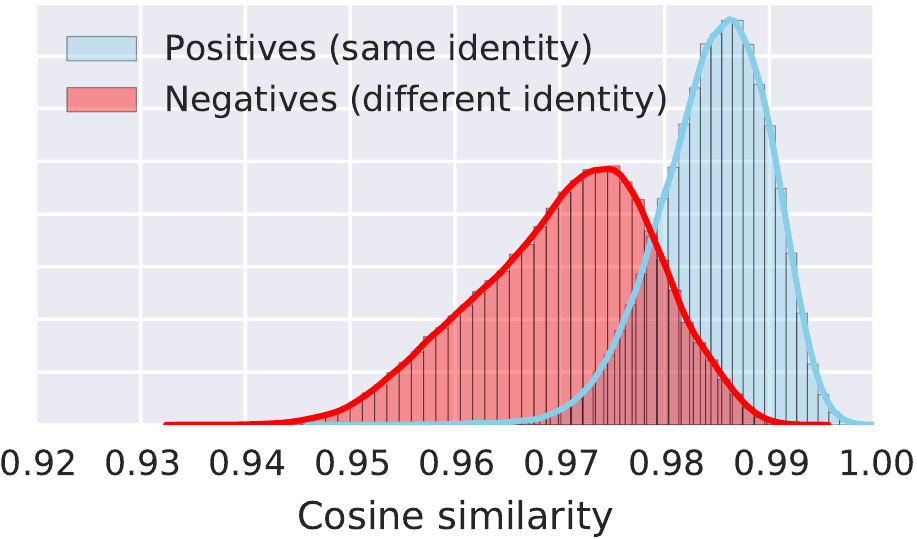}
\caption{Histograms of pairwise cosine similarity between tracks of same identity (pos) and different identity (neg) for BBT-0101.}
\label{fig:dist_histogram}
\end{figure}

\vspace{2mm}
\noindent\textbf{Role of effective mining of +/- pairs.} We emphasize that especially in light of CNN face representations, the features are very similar even across different identities, and thus positive and negative pairs need to be created properly to make the most out of them. Fig.~\ref{fig:dist_histogram} proves this point as we see a large overlap between the cosine similarity distributions of positive (same id) and negative (across id) track pairs on the base features. 

\vspace{2mm}
\noindent\textbf{TSiam, impact of singleton tracks.}
Previous work with video-level constraints~\cite{imptriplet,jfac} and~\cite{bestfgr2018}, ignore singleton (not co-occurring) tracks. In TSiam, we include negative pairs for singletons based on track distances. Table~\ref{table:tsiam_negs} shows that 30-70\% tracks are singleton and ignoring them lowers accuracy by 4\%. This confirms our hypothesis that incorporating negative pairs of singletons helps improve performance.

\begin{table}[h]
\tabcolsep=0.15cm
\small
\begin{center}
\caption{Ignoring singleton tracks (and possibly characters) leads to significant performance drop.
Accuracy on track-level clustering.}
\label{table:tsiam_negs}
\begin{tabular}{lccccc}
\toprule
			& \multicolumn{2}{c}{TSiam}		& \multicolumn{3}{c}{\# Tracks} \\
Dataset		& w/o Single~\cite{bestfgr2018} & Ours & Total & Single & Co-oc \\
\midrule
BBT-0101 	& 0.936& \textbf{0.964} & 644 & 219 & 425 \\
BF-0502 	& 0.849& \textbf{0.893} & 568 & 345 & 223 \\
\bottomrule
\end{tabular}
\end{center}
\end{table} 

\vspace{2mm}
\noindent\textbf{SSiam, comparison to pseudo-RF~\cite{psuedo1,pseudo2}.}
In Pseudo-RF, all samples are treated independent of each other, there is no batch of data from which $2K$ pairs are chosen.
A pair of samples closest in distance are chosen as positive, and farthest as negative.
However, this usually corresponds to samples that already satisfy the loss margin, thus leading to small (possibly even 0) gradient updates.
Table~\ref{table:pseudorf} shows that SSiam that involves sorting a batch of queries is much more efficient over pseudo-RF as it has the potential to select harder positives and negatives.
We see a consistent gain in performance, 3\% for BBT-0101 and over 9\% for BF-0502.

\begin{table}[h]
\small
\tabcolsep=0.3cm
\begin{center}
\caption{Comparison of \emph{SSiam} with \emph{pseudo-RF}.}
\label{table:pseudorf} 
\begin{tabular}{lccc}
\toprule 
Method			& BBT-0101 	    &  BF-0502 	   \\
\midrule
Pseudo-RF    	& 0.930 		& 0.814		  \\
SSiam     		& \textbf{0.962} 		& \textbf{0.909}	  \\
\bottomrule
\end{tabular}
\end{center}
\end{table}

\begin{table}[h]
\small
\tabcolsep=0.3cm
\begin{center}
\caption{Clustering accuracy computed at track-level on the training episodes, with a comparison to all evaluated models.}
\label{table:unsupervised_training} 
\begin{tabular}{lccccc}
\toprule 
Train/Test 	& Base  	& TSiam     &  SSiam	   \\
\midrule
BBT-0101    & 0.932 	& \textbf{0.964} 	& 0.962		  \\
BF-0502     & 0.836 	& 0.893 	& \textbf{0.909}	  \\
\bottomrule
\end{tabular}
\end{center}
\end{table}

\vspace{2mm}
\noindent\textbf{Performance on training videos.}
We report the clustering performance on training videos in Table~\ref{table:unsupervised_training}.
Note that both TSiam and SSiam are trained in an unsupervised manner, or with automatically generated labels. Both of the proposed models SSiam and TSiam provide a large performance boost over the base VGG2 features on BBT and BF.

\begin{table}[t]
\small
\tabcolsep=0.18cm
\begin{center}
\caption{Clustering accuracy computed at track-level across episodes within the same TV series.
Numbers are averaged across 5 test episodes.}
\label{table:generalization_same_series} 
\begin{tabular}{llcccccc}
\toprule
Train 	& Test  	    & Base	& TSiam  & SSiam 	\\
\midrule
BBT-0101 & BBT-01[02-06]  & 0.935 & 0.930  & 0.914 \\
BF-0502  & BF-05[01,03-06]& 0.892	& 0.889  & 0.904 \\
\bottomrule
\end{tabular}
\end{center}
\end{table}

\vspace{2mm}
\noindent\textbf{Generalization within series.} 
In this experiment, we evaluate the generalization capability of our models.
We train on one episode each of BBT-0101 and BF-0502 and evaluate on all other episodes of the same TV series.
Table~\ref{table:generalization_same_series} reports clustering accuracy.
Both SSiam or TSiam perform similar (slightly lower/higher) to the base features, possibly due to overfitting.

\begin{table}[h]
{\small
\tabcolsep=0.3cm
\begin{center}
\caption{Clustering accuracy when evaluating across video series.
Each row indicates that the model was trained on one episode of BBT / BF, but evaluated on all 6 episodes of the two series.}
\label{table:generalization_across_series}
\resizebox{.48\textwidth}{!}{
\begin{tabular}{llcc}
\toprule 
   & Trained & BBT-01[01-06] & BF-05[01-06]  \\ 
\midrule
\multirow{2}{*}{TSiam}
    & BBT-0101    & 0.936 & 0.875  \\ 
    & BF-0502     & 0.915 & 0.890  \\ 
\midrule
\multirow{2}{*}{SSiam}
    & BBT-0101    & 0.922 & 0.862 \\ 
    & BF-0502     & 0.883 & 0.905  \\ 
\bottomrule
\end{tabular}}
\end{center}}
\end{table}

\vspace{2mm}
\noindent\textbf{Generalization across series.}
We further analyze our models by evaluating generalization across series.
From Table~\ref{table:generalization_across_series}, we wish to point out an interesting observation:
TSiam and SSiam retain their discriminative power and can transfer to other series gently.
As they learn to score similarity between pairs of faces, the actual identity and characters do not seem to matter much.
For example, the drop when training TSiam on BBT-0101 and evaluating on BF is 0.890 (train on BF-0502) to 0.875.

Please note that the generalization experiments are presented here to explore the underlying properties of our models.
If achieving high performance is the only goal, we assert that our models can be trained and evaluated on each video rapidly and fully automatically.


\begin{table}[h]
{\small
\tabcolsep=0.14cm
\caption{Clustering accuracy when extending to all named characters within the episode.
BBT-0101 has 5 main and 6 named characters.
BF-0502 has 6 main and 12 named characters.}
\label{table:generalization_more_characters} 
\begin{center}
\begin{tabular}{l|cc|cc}
\toprule
    & \multicolumn{2}{c}{BBT-0101}  & \multicolumn{2}{|c}{BF-0502} \\
     	 & TSiam & SSiam        	  & TSiam & SSiam      \\
\midrule
Main cast  		& 0.964 & 0.922      	 & 0.893 & 0.905     \\
All named cast  & 0.958 & 0.922  		 & 0.829 & 0.870      \\
\bottomrule
\end{tabular}
\end{center}}
\end{table}

\vspace{2mm}
\noindent\textbf{Generalization to unseen characters.}
In the ideal setting, we would like to cluster all characters appearing in an episode including (main, other named, unknown, and background).
However, this is a very difficult setting, and in fact, disambiguating background characters is even hard for humans and there are no datasets that include such labels.
For BBT and BF, we do however have all named characters labeled.
Firstly, expanding the clustering experiment to include them drastically changes the class balance.
For example, BF-0502 has 6 main and 12 named characters with class balance 36.2/5.0 to 40.8/0.1.

We present clustering accuracy for this setting in Table~\ref{table:generalization_more_characters}.
Both proposed methods show a drop in performance when extending to unseen characters. Note that the models have been trained on only the main characters data and tested on all (including unseen) characters. 
However, the drop is small when adding just 1 new character (BBT-0101) vs. introduction of 6 in BF-0502.

SSiam's performance scales well, probably since it is trained with a diverse set of pairs (dynamically generated during training) and can generalize to unseen characters.

\subsection{Comparison with the state-of-the-art}

\noindent\textbf{BBT and BF.}
We compare our proposed methods (TSiam, and SSiam) with the state-of-the-art approaches in Table~\ref{table:soa}.
We report clustering accuracy (\%) on two videos: BBT-0101 and BF-0502.
Historically, previous works have reported performance at a frame-level. We follow this for TSiam and SSiam.

Note that our evaluation uses 2-4 times larger number of frames than previous works~\cite{jfac,imptriplet} making direct comparison hard. Specifically in BBT-0101 we have 41,220 frames while \cite{imptriplet} uses 11,525 frames. Similarly, we use 39,263 frames for BF-0502 (vs. 17,337~\cite{jfac}).
Even though we cluster more frames and tracks (with more visual diversity), our approaches are comparable to or even better than the current results.

TSiam and SSiam are better than the improved triplet method~\cite{imptriplet} on BBT-0101.
SSiam obtains 99.04\% accuracy which is 3.04\% higher performance in absolute gains.
On BF-0502, TSiam performs the best with 92.46\% which is 0.33\% better than the JFAC~\cite{jfac}.

\begin{table}[t]
\tabcolsep=0.15cm
\begin{center}
\caption{Comparison to state-of-the-art.
Metric is clustering accuracy (\%) evaluated at frame level.
Please note that many previous works use fewer tracks (\# of frames) (also indicated in Table~\ref{table:stats}) making the task relatively easier.
We use an updated version of face tracks provided by~\cite{baeuml2013}.}
\label{table:soa}
\resizebox{.48\textwidth}{!}{
\begin{tabular}{lcc|cc}
\toprule
\multirow{2}{*}{Method} &
\multirow{2}{*}{BBT-0101} & \multirow{2}{*}{BF-0502}  &
						 \multicolumn{2}{c}{Data Source} \\
                          & & &  BBT & BF \\
\midrule
\midrule
ULDML {\scriptsize(ICCV '11)}~\cite{cinbis2011unsupervised}  & 57.00  & 41.62    & $-$ & \cite{cinbis2011unsupervised} \\
HMRF {\scriptsize(CVPR '13)}~\cite{wu2013constrained}        & 59.61   & 50.30    & \cite{roth2012robust} & \cite{everingham2006}  \\
HMRF2 {\scriptsize(ICCV '13)}~\cite{wu2013simultaneous} 	 &  66.77  & $-$      & \cite{roth2012robust} & $-$ \\
WBSLRR {\scriptsize(ECCV '14)}~\cite{xiao2014weighted}       & 72.00  & 62.76    & $-$ & \cite{everingham2006} \\
VDF {\scriptsize(CVPR '17)}~\cite{sharma2017}       		& 89.62  & 87.46     & \cite{baeuml2013} &\cite{baeuml2013}\\
Imp-Triplet {\scriptsize(PacRim '16)}~\cite{imptriplet}      & 96.00  & $-$       & \cite{roth2012robust}  & $-$\\
JFAC {\scriptsize(ECCV '16)}~\cite{jfac}                     & $-$    & 92.13    & $-$ & \cite{everingham2006} \\
\midrule
\multicolumn{5}{c}{Ours (with HAC)} \\
\midrule
TSiam                                       & \textbf{98.58} & \textbf{92.46}  & &  \\
SSiam                                       & \textbf{99.04} & 90.87  & \cite{baeuml2013}*   & \cite{baeuml2013}* \\
\bottomrule
\end{tabular}}
\end{center}
\end{table}

\vspace{2mm}
\noindent\textbf{ACCIO.}
We evaluate our methods on ACCIO dataset with 36 named characters, 3243 tracks, and 166885 faces. The largest to smallest cluster ratios are very skewed: 30.65\% and 0.06\%.
In fact, half the characters correspond to less than 10\% of all tracks. Table~\ref{table:accio1} presents the results when performing clustering to yield 36 clusters (equivalent to the number of characters).
In addition, as in~\cite{jfac}, Table~\ref{table:accio2}  (num. clusters = 40) shows that our discriminative methods are not affected much by this skew, and in fact improve performance by a significant margin over the state-of-the-art.

\begin{table}[htb]
\tabcolsep=0.18cm
\begin{center}
\caption{Performance comparison of TSiam and SSiam with JFAC~\cite{jfac} on ACCIO. }
\label{table:accio1}
\resizebox{.40\textwidth}{!}{
\begin{tabular}{l|ccc}
\toprule
 & &  \#cluster=36  \\
Methods & P & R & F   \\
\midrule
JFAC {\scriptsize(ECCV '16)}~\cite{jfac}    & 0.690 & 0.350  & 0.460\\
\midrule
\multicolumn{4}{c}{Ours (with HAC)} \\
\midrule
TSiam                                       & \textbf{0.749} &  \textbf{0.382} & \textbf{0.506}\\
SSiam                                       & \textbf{0.766} & \textbf{0.386} & \textbf{0.514}\\
\bottomrule
\end{tabular}}
\end{center}
\end{table}

\begin{table}[htb]
\tabcolsep=0.15cm
\begin{center}
\caption{Performance comparison of different methods on the ACCIO dataset.}
\label{table:accio2}
\resizebox{.48\textwidth}{!}{
\begin{tabular}{l|ccc}
\toprule
 & & \# clusters=40  \\
Methods & P & R & F   \\
\midrule
DIFFRAC-DeepID2$^{+}$ {\scriptsize(ICCV '11)}~\cite{jfac}  		& 0.557  & 0.213  & 0.301 \\
WBSLRR-DeepID2$^{+}$ {\scriptsize(ECCV '14)}~\cite{jfac}       	& 0.502  & 0.206  & 0.292 \\
HMRF-DeepID2$^{+}$ {\scriptsize(CVPR '13)}~\cite{jfac}        	& 0.599  & 0.23.0  & 0.332 \\
JFAC {\scriptsize(ECCV '16)}~\cite{jfac}                     	& 0.711  & 0.352  & 0.471 \\
\midrule
\multicolumn{4}{c}{Ours (with HAC)} \\
\midrule
TSiam                                       					& \textbf{0.763} &  \textbf{0.362} & \textbf{0.491} \\
SSiam                                       					& \textbf{0.777} &  \textbf{0.371} & \textbf{0.502} \\
\bottomrule
\end{tabular}}
\end{center}
\end{table}

\vspace{2mm}
\noindent\textbf{Computational complexity.}
Our models essentially consist of a few linear layers and are very fast to compute at inference time.
In fact, training the SSiam for about 15 epochs on BBT-0101 requires less than 25 minutes (on a GTX 1080 GPU using the matconvnet framework~\cite{matconvnet}).

%% file: conclusion.tex
\section{Conclusion}
\label{sec:conclusion}

We proposed simple, unsupervised approaches for face clustering in videos, by distilling the identity factor from deep face representations.
We showed that discriminative models can leverage dynamic generation of positive/negative constraints based on ordered face distances and do not have to only rely on track-level information that is typically used.
Our proposed models are unsupervised (or use automatically generated labels) and can be trained and evaluated efficiently as they involve only a few matrix multiplications.

We conducted experiments on three challenging video datasets, comparing their differences in usage in past works. Overall, our models are very fast to train and evaluate and outperform the state-of-the-art while operating on datasets that contain more tracks with large changes in appearance.